\documentclass[letterpaper, 10 pt, conference]{ieeeconf}  % Comment this line out if you need a4paper

\IEEEoverridecommandlockouts                              % This command is only needed if 
\usepackage[nostyle]{koporec}
\pgfplotsset{compat=1.14}
\usepackage{fontawesome}
\usepackage{csvsimple}
\usepackage{subcaption}

\RequirePackage[normalem]{ulem} %DIF PREAMBLE
\RequirePackage{color}\definecolor{RED}{rgb}{1,0,0}\definecolor{BLUE}{rgb}{0,0,1} %DIF PREAMBLE
\title{\LARGE \bf
Human-Centered Unsupervised Segmentation Fusion*
}

\author{Gregor Koporec$^{1}$ and Janez Per\v{s}$^{2}$% <-this % stops a space
\thanks{*This work was supported by Gorenje, d. o. o. and by the Slovenian Research Agency (ARRS) research project J-9433 and research program P2-0095.}% <-this stops a space
\thanks{$^{1}$Gregor Koporec is with Gorenje, d. o. o.,  SI-3320 Velenje, Slovenia
        {\tt\small gregor.koporec@gorenje.com}}%
\thanks{$^{2}$Janez Per\v{s} is with the Faculty of Electrical Engineering,
University of Ljubljana, SI-1000 Ljubljana, Slovenia
        {\tt\small janez.pers@fe.uni-lj.si}}%
}

\begin{document}

\maketitle
\thispagestyle{empty}
\pagestyle{empty}

%%%%%%%%%%%%%%%%%%%%%%%%%%%%%%%%%%%%%%%%%%%%%%%%%%%%%%%%%%%%%%%%%%%%%%%%%%%%%%%%
% !TEX root = ../main.tex
\begin{abstract}
   Segmentation is generally an ill-posed problem since it results in multiple solutions and is, therefore, hard to define ground truth data to evaluate algorithms. The problem can be naively surpassed by using only one annotator per image, but such acquisition doesn't represent the cognitive perception of an image by the majority of people. Nowadays, it is not difficult to obtain multiple segmentations with crowdsourcing, so the only problem that stays is how to get one ground truth segmentation per image. There already exist numerous algorithmic solutions, but most methods are supervised or don't consider confidence per human segmentation. In this paper, we introduce a new segmentation fusion model that is based on K-Modes clustering. Results obtained from publicly available datasets with human ground truth segmentations clearly show that our model outperforms the state-of-the-art on human segmentations.
\end{abstract}

\section{Introduction}
Image segmentation is a very important step in image analysis and is typically used to combine pixels into regions corresponding to objects or parts of objects \cite{Wang2014a}. Segmentation is generally an ill-posed problem since it results in multiple solutions \cite{Mignotte2014} and is, therefore, hard to define ground truth data to evaluate algorithms. The most illustrative example of the problem is the BSDS300 dataset \cite{Martin2001} where multiple people segmented the same images differently. Even if we assume two human annotators had the same cognitive perception of an image, they may segment it at different levels of granularity \cite{Martin2001}. 

\begin{figure}[!t]
	\centering
	
	\begin{subfigure}[b]{0.3\linewidth}
	    \includepdf{1}{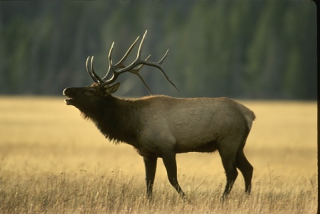}
	\end{subfigure}
	~
	\begin{subfigure}[b]{0.3\linewidth}
	    \includegraphics[width=\linewidth, frame]{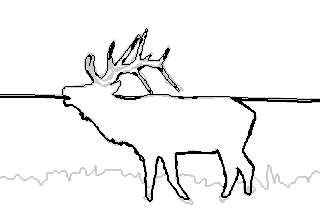}
	\end{subfigure}
	~
	\begin{subfigure}[b]{0.3\linewidth}
    	\includepdf{1}{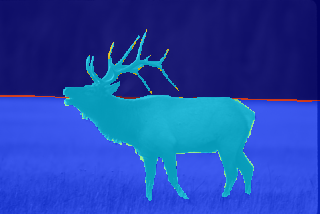}
	\end{subfigure}
	~
	\begin{subfigure}[b]{0.3\linewidth}
    	\includepdf{1}{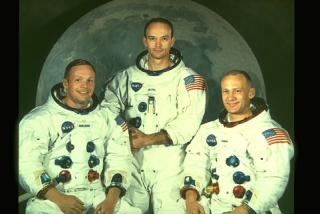}
	\end{subfigure}
	~
	\begin{subfigure}[b]{0.3\linewidth}
    	\includegraphics[width=\linewidth, frame]{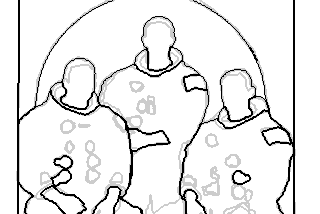}
	\end{subfigure}
	~
	\begin{subfigure}[b]{0.3\linewidth}
    	\includepdf{1}{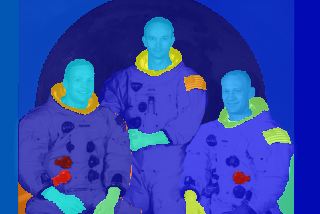}
	\end{subfigure}
	~
	\begin{subfigure}[b]{0.3\linewidth}
    	\includepdf{1}{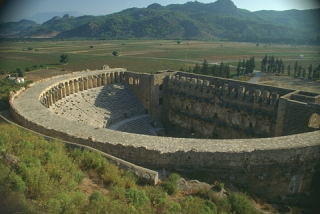}
	\end{subfigure}
	~
	\begin{subfigure}[b]{0.3\linewidth}
    	\includegraphics[width=\linewidth, frame]{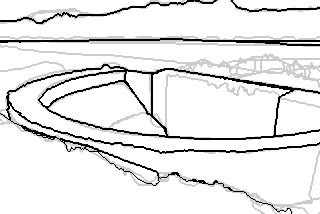}
	\end{subfigure}
	~
	\begin{subfigure}[b]{0.3\linewidth}
    	\includepdf{1}{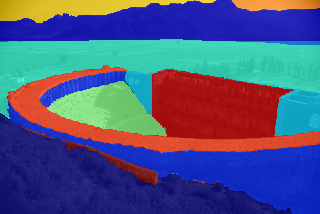}
	\end{subfigure}
	\caption{Examples of images from BSDS300 dataset \cite{Martin2001} and consensus segmentation generated by OURS VEC model.}
	\label{fig:best}
\end{figure}

The problem can be naively surpassed by using only one annotator per image when generating ground truth data and using an expert to evaluate the results as it was done in the MS-COCO dataset \cite{Lin2014}. But such acquisition of ground truth is from a statistical standpoint non-significant as the segmentation doesn't represent the cognitive perception of an image by the majority of people. Image segmentation can be viewed as a categorization of pixels and, therefore, by \cite{Lakoff1987, Wilson2004} strongly depends on the human population. So, the segmentation of one annotator is not necessarily the one that the majority of people will agree upon. 

Nowadays, it is not difficult to obtain multiple segmentations from different people because there exist multiple online crowdsourcing platforms such as Amazon Mechanical Turk or Clickworker. But using crowdsourcing platforms doesn't guarantee quality results. When obtaining data in such a way, additional confidence measures should be included. Confidence to each annotation result can be defined by human experts or automatic algorithms (\eg algorithm that determines if an object on the image was colorized beyond its borders).

The only problem that stays is how to get one ground truth segmentation per image from multiple human segmentations. Obviously, this must be a consensus segmentation that the majority of people agree upon.

\begin{figure*}[!tph]
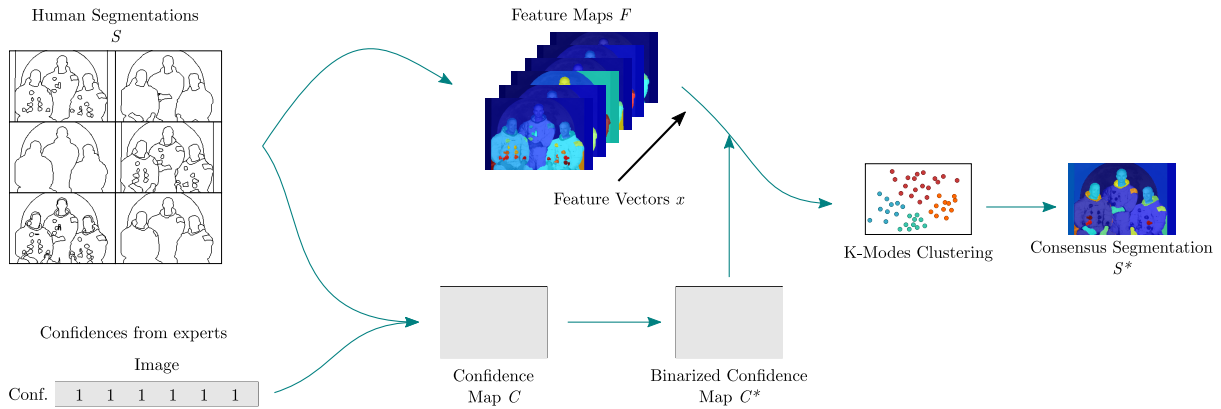

	\centering
	\medskip % for IROS19 margins
	%{\includesvg{0.9}{proposed_method}}
	{\includepdf{0.9}{proposed_method}}
	\caption{Processing pipeline of our proposed method. Human segmentations are used to get feature maps. The same segmentations with the conjunction of confidences from experts are used to get a confidence map. A binarized confidence map (a mask) is used to transform all non-significant pixels into the background region for each feature map. Masked feature maps are then used as an input set of feature vectors into K-Modes clustering. Finally, the clustering result is transformed into optimal consensus segmentation.}
	\label{fig:proposed}
\end{figure*}

There exist numerous algorithmic solutions to this problem. These are so-called segmentation fusion models. Segmentation fusion is a set of methods that are used to merge several image segmentations to get a final better segmentation \cite{Khelifi2017}. Most of the methods are supervised, where we need training data. There also exist some unsupervised methods which are mainly dependent on defining a parameter. None of the methods consider using confidences of the human segmentations.

In this paper, we introduce a new human-centered unsupervised segmentation fusion method (Figure \ref{fig:proposed}). The method is based on \mbox{K-Modes} clustering \cite{Huang1997} which sufficiently clusters categorical feature vectors into consensus segmentation regions. We also introduce a new initialization method for \mbox{K-Modes} clustering based on feature vector density. The fusion method's pipeline also includes a confidence map that is generated from expert confidences. It doesn't need parameter estimation as related methods. New fusion model is evaluated on BSDS300 \cite{Martin2001} and BSDS500 \cite{Arbelaez2011} datasets. Examples on the BSDS300 dataset are shown in Figure \ref{fig:best}. It outperforms state-of-the-art models when using human segmentations. What is more it also outperforms the average between humans (see \cite{Yang2008}).

\section{Related Work}
The first algorithms were mainly focusing on getting the segmentation by segmenting an image multiple times and then merging the results based on a defined criterion. Authors in \cite{Cho1997} used region adjacency graph (RAG) to get a set of different segmentations and also to merge them into a consensus segmentation. Mignotte \cite{Mignotte2008} generated segmentations by K-means clustering in different color spaces and then used local histograms in clustering to get the final segmentation. 

Ghosh \etal \cite{Ghosh2009} developed a more generic approach where segmentations from multiple algorithms can be used as an input to fusion algorithm. To fuse the segmentations the non-negative matrix factorization method was used.  The work \cite{Wattuya2008} also focused only on the fusion part of the segmentation algorithm and used a random walker based approach. 

Bayesian models for segmentation fusion were used in \cite{Mignotte2010, Wang2014a}. Wang \etal \cite{Wang2014a} defined segmentation fusion as a combinatorial optimization problem in terms of information theory. Consensus segmentation was generated according to a discrete distribution. The author in \cite{Mignotte2010} introduced an approach based on a Markov random field.

Many researchers also used optimization methods based on energy functions. In \cite{Mignotte2014} FMBFM model used precision-recall criterion in energy function. VOIBFM model is based on a variation of information criterion \cite{Mignotte2014a}. The likelihood energy function based on the GCE metric was introduced in \cite{Khelifi2017a} (GCEBFM model). Khelifi \etal \cite{Khelifi2017} proposed MOBFM model based on multi-objective optimization. Two criteria of segmentation were used, global consistency error (GCE) and the F-measure.

Alush \etal \cite{Alush2012} used integer linear programming to get the consensus segmentation. First, they over-segmented the image into superpixels. %Each segmentation was projected onto the created superpixel map. 
Linear programming was then applied to the set of binary merging decisions of neighboring superpixels to obtain the average segmentation. Superpixel approach was also used in \cite{Lefevre2019} where they constructed a superpixel map by intersecting the segmentations. Each superpixel was then assigned a confidence score that related to the consensus between the segmentations. %Confidence map can then by thresholded and used to define consensus segmentation.

Clustering methods were used in \cite{Franek2010}. The authors evaluated multiple ensemble clustering methods where they first generated super-pixels from the segmentations and then applied the clustering methods to get a consensus segmentation.

To produce the segmentation, contour detectors with additional transformation method can also be used. This was shown in \cite{Arbelaez2011} where a contour detector based on spectral clustering and a generic grouping algorithm were used. Grouping method was also used in \cite{Arbelaez2014} where they grouped the best combinations of multiscale regions from the image segmentation pyramid.

\section{Proposed Method}
Given a set of $L$ human segmentations $S = \{S_i\}_{i\leq L}$ with size $N \times M$ where each segmentation $S_i$ is a set of $J_i$ regions $\{R_j\}_{j \leq J_i}$ we want to get a consensus segmentation $S^*$ as shown in Figure \ref{fig:proposed}. 

Firstly, regions from all the segmentations get a unique global ID $\nu \in \mathbb{N}$. Each segmentation $S_i$ is then transformed into a feature map $F_i = \{ f_j\}_{j \leq N \times M}$. An element of a feature map $f_j$ is a pixel with a value $\nu$ of a region that it belongs to. By transforming segmentations into feature maps we implicitly define human annotations as categorical data. With this transform, we also ensure that annotations from different humans do not clash. The annotation of one human is not necessarily the same annotation of other humans. Also, annotations from different human subjects in the same position in the image do not have necessarily the same cognitive meaning.

Next, a confidence map is constructed from image segmentations and confidences from the experts. A confidence map is used to weight each segmentation's contribution to the final result by its quality. Normally, segmentations from human subjects are not the same quality. Some subjects annotate images too fast, others are superficial, or don't exactly follow the rules of the annotation task. Quality of segmentation can be acquired by experts (machines or humans) who evaluate the annotation results with the value on the interval $[0,1]$.

Each segmentation is firstly thresholded into a binary image $B_i$ where $0$ corresponds to background pixels and $1$ to pixels that were annotated by humans. With this kind of binarization, we focus only on annotated parts of the image. Confidence map is then acquired by \eqref{eq:conf_map} where each $p_i$ is a confidence defined by experts. Note that if we don't possess such data, all confidences can be $1.0$, which means that all segmentations are equally contributing to the final segmentation $S^*$.

\begin{equation}
    C = \frac{1}{L}\sum_{i\leq L}{B_i * p_i}
    \label{eq:conf_map}
\end{equation}

To use the confidence map $C$ it must first be binarized because feature maps consist of categorical data which are not real numbers. We use the threshold $0.75$ when binarizing the confidence map. Binarized confidence map $C^*$ now shows us which pixels in a feature map are significant in such a way that at least \proc{75} of human observers agree the pixel should be annotated with some label different than background. $C^*$ is thus used to transform all non-significant pixels into the background region for each feature map $F_i$. Note that in the case of BSDS datasets all confidences are $1$ and therefore all foreground pixels in binarized confidence map $C^*$ will be $1$ (high confidence).

From masked feature maps we then get a set of feature vectors $X = \{x_i\}_{i\leq N \times M}$ where every feature vector $x_i = \{a_j\}_{j\leq L}$ represents a pixel from an original image and its attributes represent region ID $\nu$ from each segmentation. Set $X$ is then used in K-Modes clustering algorithm which returns $P=\{P_i\}_{i \leq K}$ partitions of $X$. Using K-Modes clustering we need to define a number of clusters $K$ and initialization method. 

$K$ can be defined as an average of region count per segmentation as the number of regions in consensus segmentation must reflect the number of regions defined by the majority of human annotators. 

For the initialization method, the best method to date is based on attribute density \cite{Cao2009}. The attribute density method generates initial centroids by focusing on annotations of an individual human observer. But what we want are centroids on regions that the majority of human observers agree upon. In such a case, the clustering algorithm will have to cluster only the pixels where human observers don't agree if it is an element of the same region. We thus propose a new initialization method based on vector density. In this method, initial centroids are those feature vectors which are the most numerous in the set $X$.

Finally, $P$ is reshaped back to image resolution $N \times M$ and becomes $S^*$.

\section{Experimental Results}
Metric results of proposed models are shown in tables \ref{tab:bsds300} and \ref{tab:bsds500}. OURS ATTR is the model with the attribute density initialization method \cite{Cao2009} and OURS VEC is based on our feature vector density initialization method. Both methods outperform algorithms in all metrics when using human segmentations. What is more, they also outperform average human segmentation denoted as HUMAN in both datasets.

\begin{table}[!htbp]
    \caption{Region benchmarks on the BSDS300 dataset.}
	\label{tab:bsds300}
	\centering
	\begin{tabular}{
	    l 
	    S[table-format=1.2, round-mode=places, round-precision=2]
		S[table-format=1.2, round-mode=places, round-precision=2]
		S[table-format=1.2, round-mode=places, round-precision=2] 
		S[table-format=1.2, round-mode=places, round-precision=2]
	}
		\toprule
		% \faCaretDown \faChevronDown 
		\textbf{Model} & 
		\textbf{GCE \faSortAmountAsc} & 
		\textbf{VOI \faSortAmountAsc} & 
		\textbf{PRI \faSortAmountDesc} &  
		\textbf{BDE \faSortAmountAsc} \\
		\midrule
		HUMAN & 0.08024 & 1.104964 & 0.87914 & 4.882686 \\
		\midrule
		
        (Using human seg.)\\
		OURS VEC &	\boldentry{1}{2}{0.060397} & \boldentry{1}{2}{0.862702} & \boldentry{1}{2}{0.911328} & \boldentry{1}{2}{3.478571} \\
		OURS ATTR & 0.069782 &	0.908532 &	0.907478 &	3.777417 \\
		AMUS\cite{Alush2012} & 0.081 & 0.994 & 0.881 & \\
		\midrule
		(Fully automated)\\
        VOIBFM \cite{Mignotte2014a} & 0.20241 & 1.8756 & 0.80591 & 9.2976 \\
        GCEBFM \cite{Khelifi2017a} & 0.1954 & 2.105 & 0.79947 & 8.7309 \\
        FMBFM \cite{Mignotte2014} & 0.20308 & 2.0054 & 0.7964 & 8.4918 \\
		\bottomrule
	\end{tabular}
\end{table}

OURS VEC is better than OURS ATTR in all metrics. This clearly shows that the proposed initialization method better defines cluster centroids for the consensus segmentation task.
 
\begin{table}[!htbp]
    \caption{Region benchmarks on the BSDS500 dataset.}
	\label{tab:bsds500}
	\centering
	\begin{tabular}{
	        l 
			S[table-format=1.2, round-mode=places, round-precision=2] 
			S[table-format=1.2, round-mode=places, round-precision=2]
			S[table-format=1.2, round-mode=places, round-precision=2]
	}
		\toprule
		\textbf{Model} & 
		\thead{1}{COV \faSortAmountDesc} & 
		\thead{1}{PRI \faSortAmountDesc} & 
		\thead{1}{VOI \faSortAmountAsc} \\
		\midrule
		HUMAN \cite{Arbelaez2011} &	0.72 &	0.88 &	1.17\\
		\midrule
		(Using human seg.)\\
		OURS VEC & \boldentry{1}{2}{0.79} & \boldentry{1}{2}{0.91} & \boldentry{1}{2}{0.89} \\
		OURS ATTR &	{0.77} & {0.91} & {0.96}\\
		\midrule
		(Fully automated)\\
		MCG \cite{Arbelaez2014}	&	0.66 &	0.86 &	1.39 \\
        gPb-owt-ucm \cite{Arbelaez2011} &	0.65 &	0.86 &	1.48\\
        Canny-owt-ucm \cite{Arbelaez2011} &	0.55 &	0.83 &	1.89\\
        Mean Shift \cite{Arbelaez2011} &	0.58 &	0.81 &	1.64\\
		\bottomrule
	\end{tabular}
\end{table}

\begin{figure}[!htbp]
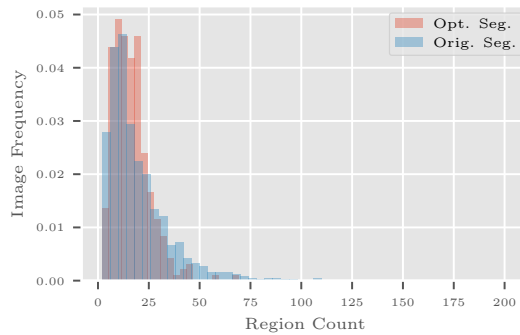

	\centering
	\includepdf{0.9}{dist_regions}
	\caption{Distributions of region count for original and consensus segmentations.}
	\label{fig:dist_regions}
\end{figure}

Using parameter $K$ as an average of region counts was a good decision. As shown in Figure \ref{fig:dist_regions} distributions of region counts for original an optimal consensus segmentations are very similar.

\begin{comment}
\begin{figure}[!htbp]
	\centering
	\includepdf{0.9}{hist_bde}
	\caption{}
	\label{fig:dist_regions}
\end{figure}

\begin{figure}[!htbp]
	\centering
	\includepdf{0.9}{hist_gce}
	\caption{}
	\label{fig:dist_regions}
\end{figure}

\begin{figure}[!htbp]
	\centering
	\includepdf{0.9}{hist_pri}
	\caption{}
	\label{fig:dist_regions}
\end{figure}

\begin{figure}[!htbp]
	\centering
	\includepdf{0.9}{hist_voi}
	\caption{}
	\label{fig:dist_regions}
\end{figure}
\end{comment}

\begin{figure}[!htbp]
    \centering
    \begin{subfigure}[b]{0.2\linewidth}
	    \includepdf{1}{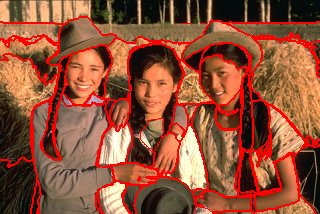}
	\end{subfigure}
	~
	\begin{subfigure}[b]{0.2\linewidth}
	    \includepdf{1}{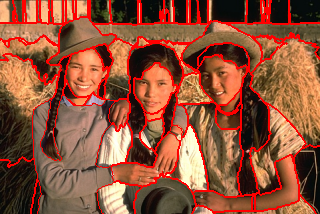}
	\end{subfigure}
	~
	\begin{subfigure}[b]{0.2\linewidth}
	    \includepdf{1}{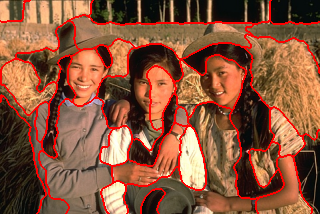}
	\end{subfigure}
	~
	\begin{subfigure}[b]{0.2\linewidth}
	    \includepdf{1}{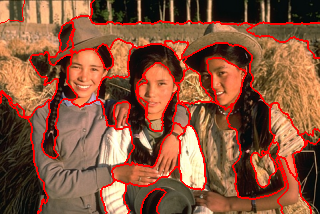}
	\end{subfigure}
	~
	\begin{comment}
	\begin{subfigure}[b]{0.2\linewidth}
	    \includepdf{1}{pic/attr_sp_54005.png}
	\end{subfigure}
	~
	\begin{subfigure}[b]{0.2\linewidth}
	    \includepdf{1}{pic/vec_sp_54005.png}
	\end{subfigure}
	~
	\begin{subfigure}[b]{0.2\linewidth}
	    \includepdf{1}{pic/voibfm_sp_54005.png}
	\end{subfigure}
	~
	\begin{subfigure}[b]{0.2\linewidth}
	    \includepdf{1}{pic/gcebfm_sp_54005.png}
	\end{subfigure}
	~
	\end{comment}
	\begin{subfigure}[b]{0.2\linewidth}
	    \includepdf{1}{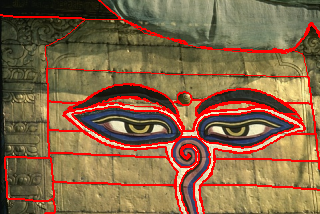}
	\end{subfigure}
	~
	\begin{subfigure}[b]{0.2\linewidth}
	    \includepdf{1}{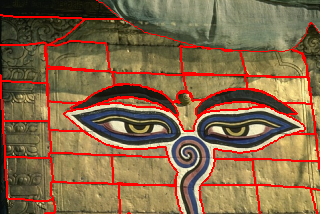}
	\end{subfigure}
	~
	\begin{subfigure}[b]{0.2\linewidth}
	    \includepdf{1}{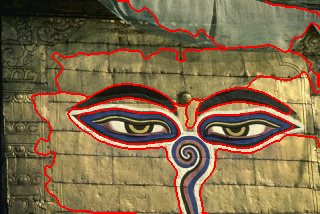}
	\end{subfigure}
	~
	\begin{subfigure}[b]{0.2\linewidth}
	    \includepdf{1}{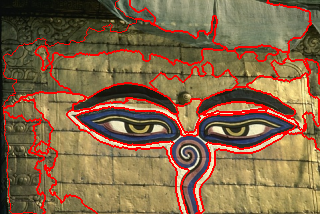}
	\end{subfigure}
	~
	\begin{subfigure}[t]{0.2\linewidth}
	    \includepdf{1}{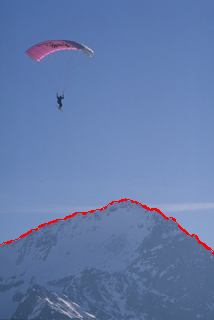}
	    \caption{OURS ATTR}
	\end{subfigure}
	~
	\begin{subfigure}[t]{0.2\linewidth}
	    \includepdf{1}{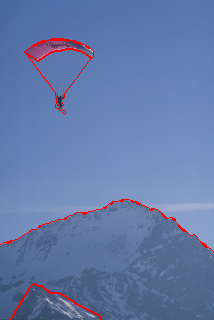}
	    \caption{OURS VEC}
	\end{subfigure}
	~
	\begin{subfigure}[t]{0.2\linewidth}
	    \includepdf{1}{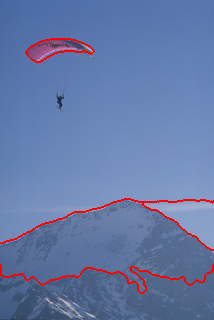}
	    \caption{VOIBFM}
	\end{subfigure}
	~
	\begin{subfigure}[t]{0.2\linewidth}
	    \includepdf{1}{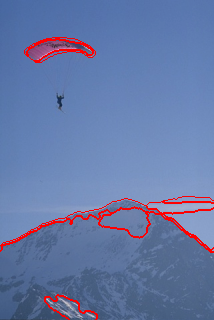}
	    \caption{GCEBFM}
	\end{subfigure}
    \caption{Examples of consensus segmentattions of different models. Images selected have the worst metric results for OURS ATTR model.}
    \label{fig:worst}
\end{figure}

Examples of consensus segmentations can be observed in Figure \ref{fig:worst}. Each column represents the results of a different algorithm. Images selected had the worst metric results for the OURS ATTR model. Such a choice was made to qualitatively assess the differences between OURS ATTR and OURS VEC. All images result in better segmentation in the case of OURS VEC. If we consider the last image with the paratrooper and two mountains, we can see that paratrooper is not selected by the OURS ATTR model. Also, a segmentation of the front mountain peak is missing. If we compare OURS VEC to VOIBFM and GCEBFM, we can see that the VOIBFM and GCEBFM segmented only the parachute and not the whole paratrooper. Also, the mountain segmentations in VOIBFM and GCEBFM are very different. Both segment the cloud on the right of the highest peak. Also, the mountains are somehow over-segmented.

\section{Conclusions}
In this paper, we introduced a new human-centered unsupervised segmentation fusion method. The method is based on K-Modes clustering which sufficiently clusters categorical feature vectors into consensus segmentation regions. We also introduced a new initialization method for K-Modes clustering based on feature vector density. The new method is more suitable for the segmentation fusion problem. The model’s pipeline also includes a confidence map that is generated from expert confidences. They can be retrieved by human or machine expert systems.

Qualitative and quantitative results show that both our models outperform state-of-the-art when using human segmentations as input by a large margin. Metrics also indicate that models outperform average human benchmarks. Consensus segmentations results show that most exposed and visible objects are correctly segmented. No excess regions are generated. The results are clear: K-Modes clustering is by far the best choice as regards segmentation fusion. Furthermore, the proposed model %does not need parameter estimation and even more, 
doesn't need training. For these models to work, we don't need thousands of training images.

What this opens up to the computer vision community is the efficient ability to automate the production of ground truth segmentation data from crowdsourcing platforms. With segmentation fusion, we can now use multiple annotators per image and get a consensus segmentation that is more objective, more statistically relevant.
%\input{sec/abstract_diff.tex}
%%%%%%%%%%%%%%%%%%%%%%%%%%%%%%%%%%%%%%%%%%%%%%%%%%%%%%%%%%%%%%%%%%%%%%%%%%%%%%%%

\addtolength{\textheight}{-12cm}   % This command serves to balance the column lengths

%%%%%%%%%%%%%%%%%%%%%%%%%%%%%%%%%%%%%%%%%%%%%%%%%%%%%%%%%%%%%%%%%%%%%%%%%%%%%%%%

%%%%%%%%%%%%%%%%%%%%%%%%%%%%%%%%%%%%%%%%%%%%%%%%%%%%%%%%%%%%%%%%%%%%%%%%%%%%%%%%

%%%%%%%%%%%%%%%%%%%%%%%%%%%%%%%%%%%%%%%%%%%%%%%%%%%%%%%%%%%%%%%%%%%%%%%%%%%%%%%%
%\section*{APPENDIX}


%\section*{ACKNOWLEDGMENT}

%The preferred spelling of the word ÒacknowledgmentÓ in America is without an ÒeÓ after the ÒgÓ. Avoid the stilted expression, ÒOne of us (R. B. G.) thanks . . .Ó  Instead, try ÒR. B. G. thanksÓ. Put sponsor acknowledgments in the unnumbered footnote on the first page.

%%%%%%%%%%%%%%%%%%%%%%%%%%%%%%%%%%%%%%%%%%%%%%%%%%%%%%%%%%%%%%%%%%%%%%%%%%%%%%%%
\bibliographystyle{./IEEEtran}
\bibliography{main}

\begin{thebibliography}{10}
\providecommand{\url}[1]{#1}
\csname url@rmstyle\endcsname
\providecommand{\newblock}{\relax}
\providecommand{\bibinfo}[2]{#2}
\providecommand\BIBentrySTDinterwordspacing{\spaceskip=0pt\relax}
\providecommand\BIBentryALTinterwordstretchfactor{4}
\providecommand\BIBentryALTinterwordspacing{\spaceskip=\fontdimen2\font plus
\BIBentryALTinterwordstretchfactor\fontdimen3\font minus
  \fontdimen4\font\relax}
\providecommand\BIBforeignlanguage[2]{{%
\expandafter\ifx\csname l@#1\endcsname\relax
\typeout{** WARNING: IEEEtran.bst: No hyphenation pattern has been}%
\typeout{** loaded for the language `#1'. Using the pattern for}%
\typeout{** the default language instead.}%
\else
\language=\csname l@#1\endcsname
\fi
#2}}

\bibitem{Wang2014a}
H.~Wang, Y.~Zhang, R.~Nie, Y.~Yang, B.~Peng, and T.~Li, ``Bayesian image
  segmentation fusion,'' \emph{Knowledge-Based Systems}, vol.~71, pp. 162--168,
  2014.

\bibitem{Mignotte2014}
M.~Mignotte and C.~Hélou, ``A precision-recall criterion based consensus model
  for fusing multiple segmentations,'' \emph{International Journal of Signal
  Processing, Image Processing and Pattern Recognition}, vol.~7, no.~3, pp.
  61--82, 2014.

\bibitem{Martin2001}
D.~Martin, C.~Fowlkes, D.~Tal, and J.~Malik, ``A database of human segmented
  natural images and its application to evaluating segmentation algorithms and
  measuring ecological statistics.''\hskip 1em plus 0.5em minus 0.4em\relax
  Iccv Vancouver:, 2001.

\bibitem{Lin2014}
T.-Y. Lin, M.~Maire, S.~Belongie, L.~Bourdev, R.~Girshick, J.~Hays, P.~Perona,
  D.~Ramanan, C.~L. Zitnick, and P.~Dol{\'{i}}, ``{Microsoft COCO: Common
  Objects in Context},'' in \emph{Computer Vision -- ECCV 2014}, 2014, pp.
  740--755.

\bibitem{Lakoff1987}
G.~Lakoff, \emph{{Women, fire, and dangerous things}}.\hskip 1em plus 0.5em
  minus 0.4em\relax Chicago: The University of Chicago Press, 1987.

\bibitem{Wilson2004}
D.~Wilson and D.~Sperber, ``{Relevance Theory},'' in \emph{The Handbook of
  Pragmatics}, 2004, ch.~27, pp. 607--632.

\bibitem{Khelifi2017}
L.~Khelifi and M.~Mignotte, ``A multi-objective decision making approach for
  solving the image segmentation fusion problem,'' \emph{IEEE Transactions on
  Image Processing}, vol.~26, no.~8, pp. 3831--3845, 2017.

\bibitem{Huang1997}
Z.~Huang, ``{Clustering large data sets with mixed numeric and categorical
  values},'' in \emph{Proceedings of the First Pacific Asia Knowledge Discovery
  and Data Mining Conference}, 1997, pp. 21--34.

\bibitem{Arbelaez2011}
P.~Arbelaez, M.~Maire, C.~Fowlkes, and J.~Malik, ``Contour detection and
  hierarchical image segmentation,'' \emph{IEEE Trans. Pattern Anal. Mach.
  Intell.}, vol.~33, no.~5, pp. 898--916, May 2011.

\bibitem{Yang2008}
A.~Y. Yang, J.~Wright, Y.~Ma, and S.~S. Sastry, ``Unsupervised segmentation of
  natural images via lossy data compression,'' \emph{Computer Vision and Image
  Understanding}, vol. 110, no.~2, pp. 212--225, 2008.

\bibitem{Cho1997}
K.~Cho and P.~Meer, ``Image segmentation from consensus information,''
  \emph{Computer Vision and Image Understanding}, vol.~68, no.~1, pp. 72--89,
  1997.

\bibitem{Mignotte2008}
M.~Mignotte, ``Segmentation by fusion of histogram-based$k$-means clusters in
  different color spaces,'' \emph{IEEE Transactions on Image Processing},
  vol.~17, pp. 780--787, 2008.

\bibitem{Ghosh2009}
S.~Ghosh, J.~J. Pfeiffer, and J.~Mulligan, ``A general framework for
  reconciling multiple weak segmentations of an image,'' in \emph{2009 Workshop
  on Applications of Computer Vision (WACV)}.\hskip 1em plus 0.5em minus
  0.4em\relax IEEE, 2009, pp. 1--8.

\bibitem{Wattuya2008}
P.~Wattuya, K.~Rothaus, J.-S. Prassni, and X.~Jiang, ``A random walker based
  approach to combining multiple segmentations.''\hskip 1em plus 0.5em minus
  0.4em\relax IEEE, 2008, pp. 1--4.

\bibitem{Mignotte2010}
M.~Mignotte, ``A label field fusion bayesian model and its penalized maximum
  rand estimator for image segmentation,'' \emph{IEEE Transactions on Image
  Processing}, vol.~19, no.~6, pp. 1610--1624, 2010.

\bibitem{Mignotte2014a}
M.~{Mignotte}, ``A label field fusion model with a variation of information
  estimator for image segmentation,'' \emph{Information Fusion}, vol.~20, pp.
  7--20, 2014.

\bibitem{Khelifi2017a}
L.~Khelifi and M.~Mignotte, ``A novel fusion approach based on the global
  consistency criterion to fusing multiple segmentations,'' \emph{IEEE
  Transactions on Systems, Man, and Cybernetics: Systems}, vol.~47, no.~9, pp.
  2489--2502, 2017.

\bibitem{Alush2012}
A.~{Alush} and J.~{Goldberger}, ``Ensemble segmentation using efficient integer
  linear programming,'' \emph{IEEE Transactions on Pattern Analysis and Machine
  Intelligence}, vol.~34, no.~10, pp. 1966--1977, 2012.

\bibitem{Lefevre2019}
S.~Lef{\`e}vre, D.~Sheeren, and O.~Tasar, ``A generic framework for combining
  multiple segmentations in geographic object-based image analysis,''
  \emph{ISPRS International Journal of Geo-Information}, vol.~8, no.~2, p.~70,
  2019.

\bibitem{Franek2010}
L.~Franek, D.~D. Abdala, S.~Vega-Pons, and X.~Jiang, ``Image segmentation
  fusion using general ensemble clustering methods,'' in \emph{Asian Conference
  on Computer Vision}.\hskip 1em plus 0.5em minus 0.4em\relax Springer, 2010,
  pp. 373--384.

\bibitem{Arbelaez2014}
P.~Arbel\'{a}ez, J.~Pont-Tuset, J.~Barron, F.~Marques, and J.~Malik,
  ``Multiscale combinatorial grouping,'' in \emph{Computer Vision and Pattern
  Recognition}, 2014.

\bibitem{Cao2009}
F.~Cao, J.~Liang, and L.~Bai, ``{A new initialization method for categorical
  data clustering},'' \emph{Expert Systems with Applications}, vol.~36, no.~7,
  pp. 10\,223--10\,228, 2009.

\end{thebibliography}

\end{document}